\relax
\documentclass[letterpaper]{article} 
\usepackage{aaai21}  
\usepackage{times}  
\usepackage{helvet} 
\usepackage{courier}  
\usepackage[hyphens]{url}  
\usepackage{graphicx} 
\usepackage{natbib}  
\usepackage{caption} 
\newcommand{\eg}{{\textit{e.g.}}}
\newcommand{\ie}{{\textit{i.e.}}}
\newcommand{\etal}{{\textit{et al.}}}

\usepackage{graphicx}
\usepackage{xtab,longtable}
\usepackage{multirow}
\usepackage{amsmath}
\usepackage{makecell}
\usepackage{tabularx}
\usepackage[title,toc,titletoc,page]{appendix}
\usepackage[caption=false]{subfig}
\usepackage{comment}
\usepackage{booktabs}
\usepackage{array}
\newcommand{\PreserveBackslash}[1]{\let\temp=\\#1\let\\=\temp}
\newcolumntype{C}[1]{>{\PreserveBackslash\centering}p{#1}}

\title{An Experimental Study of Weight Initialization and Weight Inheritance Effects on Neuroevolution\thanks{This material is based upon work supported by the U.S. Department of Energy, Office of Science, Office of Advanced Combustion Systems under Award Number \#FE0031547}}
\author{
    Zimeng Lyu,
    AbdElRahman ElSaid,
    Joshua Karns,
    Mohamed Mkaouer,
    Travis Desell
    \\
}
\affiliations{

      Rochester Institute of Technology\\
      Rochester, NY 14623 \\
    zimenglyu@mail.rit.edu, aelsaid@mail.rit.edu, josh@mail.rit.edu, mwmvse@rit.edu, tjdvse@rit.edu

}

\begin{document}

\maketitle

\begin{abstract}
Weight initialization is critical in being able to successfully train artificial neural networks (ANNs), and even more so for recurrent neural networks (RNNs) which can easily suffer from vanishing and exploding gradients. In neuroevolution, where evolutionary algorithms are applied to neural architecture search, weights typically need to be initialized at three different times: when initial genomes (ANN architectures) are created at the beginning of the search, when offspring genomes are generated by crossover, and when new nodes or edges are created during mutation. This work explores the difference between using Xavier, Kaiming, and uniform random weight initialization methods, as well as novel Lamarckian weight inheritance methods for initializing new weights during crossover and mutation operations. These are examined using the Evolutionary eXploration of Augmenting Memory Models (EXAMM) neuroevolution algorithm, which is capable of evolving RNNs with a variety of modern memory cells (e.g., LSTM, GRU, MGU, UGRNN and Delta-RNN cells) as well recurrent connections with varying time skips through a high performance island based distributed evolutionary algorithm. Results show that with statistical significance, utilizing the Lamarckian strategies outperforms Kaiming, Xavier and uniform random weight initialization, and can speed neuroevolution by requiring less backpropagation epochs to be evaluated for each generated RNN.

\end{abstract}

\section{Introduction}


Neuroevolution, or the use of evolutionary algorithms for neural architecture search and training, has seen a significant growth in popularity and wide use due to the challenges of designing deep neural networks~\cite{stanley2019designing,liu2020survey}. While there are some approaches to neuroevolution such as indirect encoding, \eg, HyperNEAT~\cite{stanley2009hypercube}, where the genomes are used to generate the architecture and assign weights, or fitness estimation, \eg,~\cite{camero2019specialized,camero2018low} where neural network fitness is estimated without training the networks, most modern neuroevolution algorithms involve a direct encoding approach, where a neural network's architecture and weights are directly represented as \emph{genomes} which can be evolved by crossover and mutation operations.

For these direct encoding strategies, how the network weights are initialized is critical, especially for deep neural networks (DNNs)~\cite{pascanu2013difficulty} as it has been shown that poor weight initialization quickly leads to gradient vanishing and exploding problems~\cite{glorot2010understanding}. The Xavier~\cite{glorot2010understanding}~and Kaiming weight initialization~\cite{he2015delving}~ methods have been a great success in reducing issues for DNNs and are now the de facto standard for training DNNs, however these methods do not take into account extra information available during neuroevolution. For example, during mutation a child genome is generated by randomly modifying a previously trained parent genome, and during crossover a child genome is generated utilizing two (or more) previously trained parental genomes. These parental weight values and distributions contain valuable information which can be used to better initialize child genome weights, and this process is known as \emph{Lamarckian}~\cite{real2017large, prellberg2018lamarckian, deb2002fast} or sometimes \emph{Epigenetic}~\cite{desell2018accelerating} weight initialization.

Unfortunately, many neuroevolution algorithms still use an outdated uniform random initialization for initial populations~\cite{stanley2002evolving, zhang2007moea, elsaid2018optimizing, ororbia2019examm}, with with a few exceptions that use Xavier~\cite{aly2019optimizing,prellberg2018lamarckian} or Kaiming~\cite{desell2018accelerating} initialization. Some studies suggest that Lamarckian weight initialization can reduce the number of backpropagation (BP) epochs to train neural networks~\cite{desell2018accelerating,ku1997exploring} and lead to better performing neural networks, to the authors' knowledge Lamarckian weight inheritance has not been rigorously compared to the modern Xavier and Kaiming weight initialization methods.

This work provides an experimental analysis of Lamarckian weight inheritance methods (one for crossover and another for mutation) to Xavier, Kaiming and uniform random initialization as a baseline. It is done in the context of evolving deep recurrent neural networks (RNNs) for time series data prediction using two challenging real world data sets. Results are promising, showing that with statistical significance the Lamarckian strategies outperform Xavier and Kaiming weight initialization, and further can reduce the amount BP epochs used to train the neural networks, allowing more time to be spent on architectural evolution.

\section{Weight Initialization and Inheritance}


Xavier weight initialization~\cite{glorot2010understanding} was designed for DNNs with symmetrical activation functions such as $tanh$ and $softsign$. The weights in each layer are generated using a uniform distribution:
\begin{equation}
W \sim \mathcal{U}[-\frac{\sqrt{6}}{\sqrt{f_{in} + f_{out}}}, \frac{\sqrt{6}}{\sqrt{f_{in} + f_{out}}}]
\end{equation}
where $f_{in}$ and $f_{out}$ are fan in and fan out of the layer.

Kaiming weight initialization~\cite{he2015delving} was designed for non-symmetrical activation functions such as ReLUs. The weights in each layer are generated with a normal distribution:
\begin{equation}
W \sim N(0,1) * \frac{\sqrt{2}}{f_{in}}
\end{equation}
where $f_{in}$ is the fan in of the layer.

The Lamarckian strategies investigated in this work were first introduced by Desell \etal~for neuroevolution of convolutional neural networks (CNNs) and later used for recurrent neural networks~\cite{desell2018accelerating,ororbia2019examm}. While Prellberg and Kramer also investigated Lamarckian weight inheritance for CNNs, this was simpler version where only mutated components were re-initialized randomly~\cite{prellberg2018lamarckian}.


For direct encoding neuroevolution algorithms, after the weight initialization of initial genomes (\ie, neural network architectures), new genomes are created either via crossover, where two or more parents are recombined into a child genome, or by mutation where a single parent has one or more random modifications made.

For crossover, given a more fit and less fit parent, child genome weights are initialized as follows. When the same architectural component (\eg, node, edge or layer) exists in both parents\footnote{Components are identified as being the same by having the same \emph{innovation number}, which are uniquely created by the neuroevolution process when an architectural component is added to a genome, and are inherited by children on crossover and mutation, as in the NEAT algorithm~\cite{stanley2002evolving}.}, the weights and biases for that component are generated using a stochastic line search recombining weights or biases from those in the parents' components. Given a random number $r\sim\mathcal{U}[-0.5, 1.5]$, a child’s weight $w_c$ is set to:
\begin{equation}
    w_c = r(w_{p2} - w_{p1}) + w_{p1}
\end{equation}
where $w_{p1}$ is the weight from the more fit parent, and $w_{p2}$ is the weight from the less fit parent (note the same $r$ value is used for \emph{all} child weights). This allows the child weights to be set along a gradient calculated from the weights of the two parents, allowing for informed exploration of the weight space between and around the two parents. In the case where the component only exists in one parent, the same weights and biases are copied to the child.

For mutations, new components are added to parent neural network architecture, so it is not possible to directly utilize weights from the parent. Instead, statistical information about the weight distributions of the parents can be used. Weights and biases for new components generated during mutations are instead initialized using a normal distribution around the mean $\mu_p$ and variance $\sigma^2_p$ of the parent's weights:
\begin{equation}
    W \sim N(\mu_p, \sigma^2_p)
\end{equation}
while the other weights are copied from the parent. This network-aware approach using the statistical distribution of a network's weights has also been shown in other work to speed transfer learning, lending further credence to this approach~\cite{elsaid2020improving}.

\section{Methodology}

This work utilized the Evolutionary eXploration of Augmenting Memory Models (EXAMM) neuroevolution algorithm to explore the different weight initialization and inheritance strategies. EXAMM evolves progressively larger RNNs through a series of mutation and crossover (reproduction) operations. Mutations can be edge-based: \emph{split edge}, \emph{add edge}, \emph{enable edge}, \emph{add recurrent edge}, and \emph{disable edge} operations, or work as higher-level node-based mutations: \emph{disable node}, \emph{enable node}, \emph{add node}, \emph{split node} and \emph{merge node}. The type of node to be added is selected uniformly at random from a suite of simple neurons and complex memory cells: $\Delta$-RNN units~\cite{ororbia2017diff}, gated recurrent units (GRUs)~\cite{chung2014empirical}, long short-term memory cells (LSTMs)~\cite{hochreiter1997long}, minimal gated units (MGUs)~\cite{zhou2016minimal}, and update gate RNN cells (UGRNNs)~\cite{collins2016capacity}. This allows EXAMM to select for the best performing recurrent memory units. EXAMM also allows for \emph{deep recurrent connections} which enables the RNN to directly use information beyond the previous time step. These deep recurrent connections have proven to offer significant improvements in model generalization, even yielding models that outperform state-of-the-art gated architectures~\cite{desell2019evostar-deeprecurrent}.  EXAMM has both a multithreaded implementation and an MPI implementation for distributed use on high performance computing resources. To the authors' knowledge, these capabilities are not available in other neuroevolution frameworks capable of evolving RNNs, which is the primary reason EXAMM was selected to serve as the basis of this work. Additionally, its implementations allowing use of high performance computing resources allowed the results to be gathered in a timely matter using challenging real world time series data prediction problems. We refer the reader to Ororbia~\etal~\cite{ororbia2019examm} for more details on EXAMM.

\section{Results}

\paragraph{Data Sets}

\begin{figure*}[t]
    \centering
    \begin{minipage}[t]{.48\textwidth}
        \centering
        \includegraphics[width=0.9\textwidth]{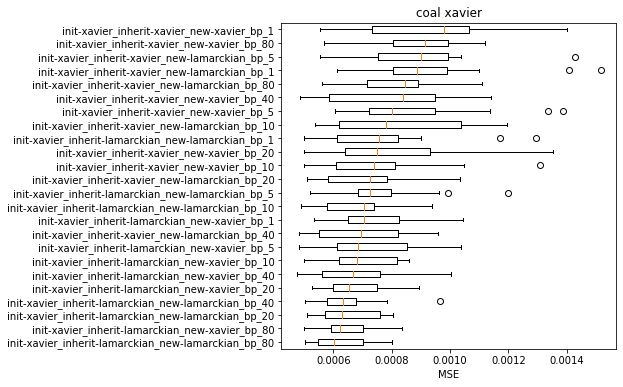}
        \includegraphics[width=0.9\textwidth]{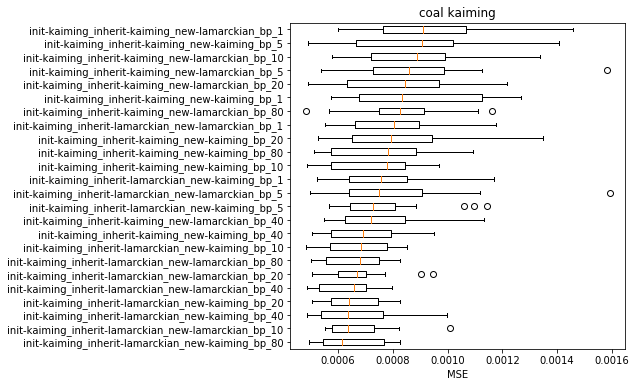}
        \includegraphics[width=0.9\textwidth]{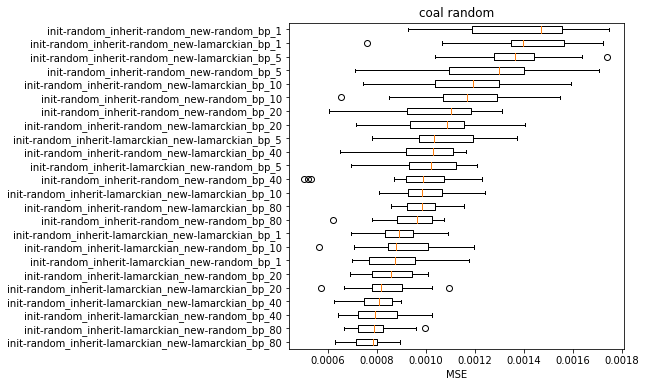}
    \end{minipage}
    \hfill
    \begin{minipage}[t]{.48\textwidth}
        \centering
        \includegraphics[width=0.9\textwidth]{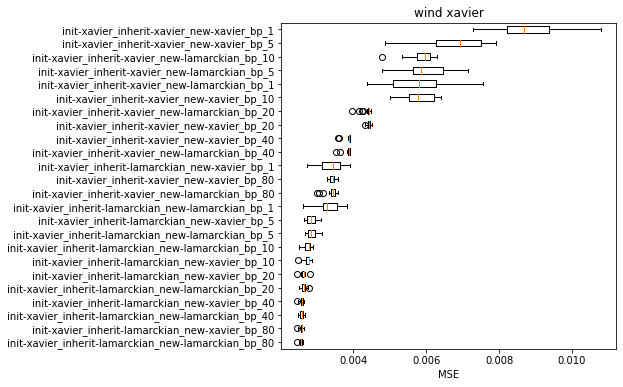}
        \includegraphics[width=0.9\textwidth]{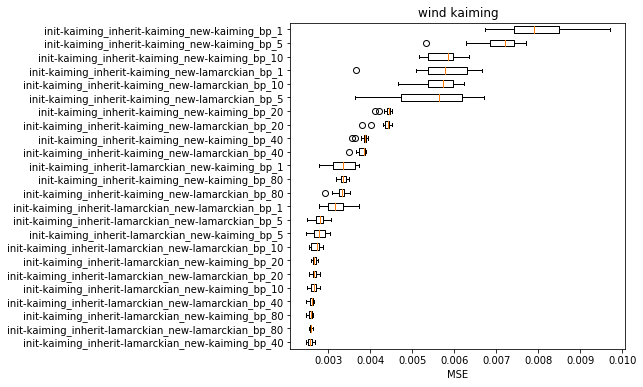}
        \includegraphics[width=0.9\textwidth]{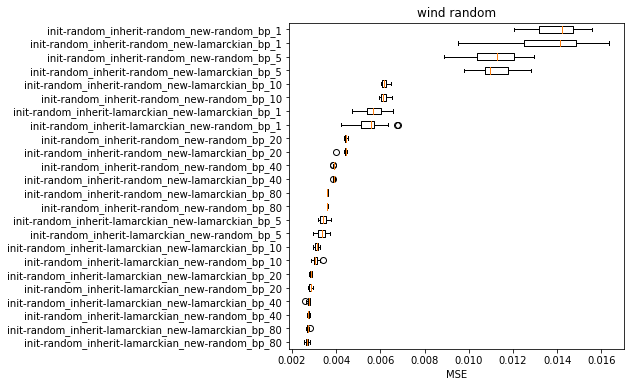}
    \end{minipage}

    \caption{\label{fig:box_plot} Convergence rates (in terms of best MSE on validation data) with Xavier, Kaiming and uniform random weight initialization predicting \emph{main flame intensity} from the coal fired power plant dataset and \emph{average active power} for the wind turbine dataset.}
\end{figure*}

    
    


\begin{table*}[p]
\centering
    \fontsize{7.5pt}{8.5pt} \selectfont
    \begin{tabular}{|C{28pt}|C{35pt}|C{48pt}|C{52pt}|C{30pt}|C{30pt}|C{30pt}|C{33pt}|C{33pt}|C{33pt}|} \hline
      BP    & Weight        & Crossover  & Mutation & Avg  & Avg   &  Avg      & Worst & Avg & Best\\
      Epochs & Initialize  & Inheritance  & New Edge/Node  & Node & Edge  & Rec Edge  & MAE   & MAE & MAE\\ \hline \hline

\multirow{12}{*}{1} & \multirow{4}{*}{Random} & Random & Random & 20.1 & 41.5 & 27.1 & 1.75e-3 & 1.38e-3 & 9.25e-4 \\ \cline{3-10}
                &                           & Lamarckian & Random & 38.1 & 50.4 & 34.3 & 1.09e-3 & 8.82e-4 & 6.96e-4 \\ \cline{3-10}
                &                           & Lamarckian & Lamarckian & 46.6 & 54.0 & 30.9 & 1.18e-3 & 8.90e-4 & 6.99e-4 \\ \cline{3-10}
                &                           & Random & Lamarckian & 23.3 & 33.5 & 46.1 & 1.72e-3 & 1.40e-3 & 7.60e-4 \\ \cline{2-10}
                & \multirow{4}{*}{Xavier} & Xavier & Xavier & 19.9 & 33.0 & 58.5 & 1.40e-3 & 9.40e-4 & 5.53e-4 \\ \cline{3-10}
                &                           & Lamarckian & Xavier & 23.9 & 36.8 & 44.8 & 1.29e-3 & 7.58e-4 & \bf4.98e-4 \\ \cline{3-10}
                &                           & Lamarckian & Lamarckian & 26.9 & 33.9 & 39.0 & \bf1.04e-3 & \bf7.39e-4 & 5.35e-4 \\ \cline{3-10}
                &                           & Xavier & Lamarckian & 18.2 & 33.0 & 48.8 & 1.52e-3 & 9.27e-4 & 6.12e-4 \\ \cline{2-10}
                & \multirow{4}{*}{Kaiming} & Kaiming & Kaiming & 19.1 & 34.1 & 40.5 & 1.27e-3 & 8.79e-4 & 5.74e-4 \\ \cline{3-10}
                &                           & Lamarckian & Kaiming & 24.4 & 46.2 & 38.7 & 1.18e-3 & 8.09e-4 & 5.53e-4 \\ \cline{3-10}
                &                           & Lamarckian & Lamarckian & 21.9 & 50.5 & 53.3 & 1.17e-3 & 7.81e-4 & 5.23e-4 \\ \cline{3-10}
                &                           & Kaiming & Lamarckian & 17.1 & 31.4 & 47.4 & 1.46e-3 & 9.36e-4 & 6.02e-4 \\ \cline{1-10}
\multirow{12}{*}{5} & \multirow{4}{*}{Random} & Random & Random & 17.0 & 35.6 & 31.5 & 1.71e-3 & 1.24e-3 & 7.12e-4 \\ \cline{3-10}
                &                           & Lamarckian & Random & 37.0 & 37.2 & 42.9 & 1.37e-3 & 1.06e-3 & 7.80e-4 \\ \cline{3-10}
                &                           & Lamarckian & Lamarckian & 31.8 & 31.1 & 50.6 & 1.21e-3 & 1.02e-3 & 6.93e-4 \\ \cline{3-10}
                &                           & Random & Lamarckian & 17.0 & 38.1 & 27.9 & 1.74e-3 & 1.36e-3 & 1.04e-3 \\ \cline{2-10}
                & \multirow{4}{*}{Xavier} & Xavier & Xavier & 16.0 & 32.2 & 24.8 & 1.39e-3 & 8.65e-4 & 6.05e-4 \\ \cline{3-10}
                &                           & Lamarckian & Xavier & 21.4 & 56.1 & 39.8 & 1.20e-3 & 7.58e-4 & 5.21e-4 \\ \cline{3-10}
                &                           & Lamarckian & Lamarckian & 21.4 & 43.8 & 39.4 & \bf1.04e-3 & \bf7.24e-4 & \bf4.83e-4 \\ \cline{3-10}
                &                           & Xavier & Lamarckian & 16.9 & 35.9 & 20.9 & 1.43e-3 & 8.87e-4 & 5.55e-4 \\ \cline{2-10}
                & \multirow{4}{*}{Kaiming} & Kaiming & Kaiming & 15.7 & 31.6 & 22.0 & 1.41e-3 & 9.03e-4 & 4.90e-4 \\ \cline{3-10}
                &                           & Lamarckian & Kaiming & 19.8 & 57.9 & 33.7 & 1.59e-3 & 8.00e-4 & 4.98e-4 \\ \cline{3-10}
                &                           & Lamarckian & Lamarckian & 21.0 & 53.5 & 42.0 & 1.14e-3 & 7.64e-4 & 5.68e-4 \\ \cline{3-10}
                &                           & Kaiming & Lamarckian & 16.1 & 33.3 & 22.9 & 1.58e-3 & 8.77e-4 & 5.38e-4 \\ \cline{1-10}
\multirow{12}{*}{10} & \multirow{4}{*}{Random} & Random & Random & 16.5 & 33.0 & 23.6 & 1.55e-3 & 1.18e-3 & 6.53e-4 \\ \cline{3-10}
                &                           & Lamarckian & Random & 38.8 & 25.8 & 47.0 & 1.24e-3 & 1.01e-3 & 8.29e-4 \\ \cline{3-10}
                &                           & Lamarckian & Lamarckian & 35.1 & 38.9 & 53.2 & 1.20e-3 & 9.15e-4 & 5.63e-4 \\ \cline{3-10}
                &                           & Random & Lamarckian & 16.1 & 32.1 & 18.1 & 1.59e-3 & 1.20e-3 & 7.42e-4 \\ \cline{2-10}
                & \multirow{4}{*}{Xavier} & Xavier & Xavier & 16.2 & 32.0 & 15.4 & 1.31e-3 & 7.51e-4 & 4.98e-4 \\ \cline{3-10}
                &                           & Lamarckian & Xavier & 19.5 & 48.6 & 30.4 & 9.38e-4 & 6.91e-4 & 4.89e-4 \\ \cline{3-10}
                &                           & Lamarckian & Lamarckian & 19.6 & 54.8 & 26.9 & 8.87e-4 & 6.99e-4 & 4.98e-4 \\ \cline{3-10}
                &                           & Xavier & Lamarckian & 16.4 & 30.7 & 17.2 & 1.19e-3 & 8.25e-4 & 5.38e-4 \\ \cline{2-10}
                & \multirow{4}{*}{Kaiming} & Kaiming & Kaiming & 16.1 & 30.6 & 15.3 & 9.70e-4 & 7.33e-4 & 4.91e-4 \\ \cline{3-10}
                &                           & Lamarckian & Kaiming & 21.9 & 51.4 & 36.8 & 1.01e-3 & \bf6.66e-4 & 5.52e-4 \\ \cline{3-10}
                &                           & Lamarckian & Lamarckian & 18.4 & 47.6 & 25.8 & \bf8.52e-4 & 6.74e-4 & \bf\emph{4.82e-4} \\ \cline{3-10}
                &                           & Kaiming & Lamarckian & 16.4 & 31.9 & 21.1 & 1.34e-3 & 8.80e-4 & 5.78e-4 \\ \cline{1-10}
\multirow{12}{*}{20} & \multirow{4}{*}{Random} & Random & Random & 16.4 & 30.1 & 15.8 & 1.31e-3 & 1.05e-3 & 6.05e-4 \\ \cline{3-10}
                &                           & Lamarckian & Random & 29.8 & 42.6 & 35.6 & 1.09e-3 & 8.37e-4 & 5.70e-4 \\ \cline{3-10}
                &                           & Lamarckian & Lamarckian & 35.0 & 46.1 & 39.4 & 1.01e-3 & 8.61e-4 & 7.19e-4 \\ \cline{3-10}
                &                           & Random & Lamarckian & 16.0 & 28.6 & 14.4 & 1.41e-3 & 1.05e-3 & 7.14e-4 \\ \cline{2-10}
                & \multirow{4}{*}{Xavier} & Xavier & Xavier & 16.4 & 29.6 & 12.0 & 1.35e-3 & 8.03e-4 & 5.00e-4 \\ \cline{3-10}
                &                           & Lamarckian & Xavier & 19.5 & 50.5 & 21.6 & \bf8.03e-4 & \bf6.48e-4 & 5.08e-4 \\ \cline{3-10}
                &                           & Lamarckian & Lamarckian & 20.1 & 50.7 & 18.4 & 8.95e-4 & 6.72e-4 & 5.27e-4 \\ \cline{3-10}
                &                           & Xavier & Lamarckian & 15.9 & 28.2 & 11.5 & 1.08e-3 & 7.28e-4 & 5.10e-4 \\ \cline{2-10}
                & \multirow{4}{*}{Kaiming} & Kaiming & Kaiming & 16.3 & 28.4 & 11.1 & 1.35e-3 & 8.12e-4 & 5.35e-4 \\ \cline{3-10}
                &                           & Lamarckian & Kaiming & 19.2 & 50.8 & 22.6 & 9.48e-4 & 6.65e-4 & 5.06e-4 \\ \cline{3-10}
                &                           & Lamarckian & Lamarckian & 18.8 & 46.0 & 20.6 & 8.27e-4 & 6.59e-4 & 5.07e-4 \\ \cline{3-10}
                &                           & Kaiming & Lamarckian & 15.7 & 26.7 & 10.6 & 1.22e-3 & 8.23e-4 & \bf4.90e-4 \\ \cline{1-10}
\multirow{12}{*}{40} & \multirow{4}{*}{Random} & Random & Random & 16.2 & 28.4 & 8.8 & 1.23e-3 & 9.53e-4 & 5.04e-4 \\ \cline{3-10}
                &                           & Lamarckian & Random & 24.4 & 48.8 & 20.1 & 8.98e-4 & 7.97e-4 & 6.26e-4 \\ \cline{3-10}
                &                           & Lamarckian & Lamarckian & 23.6 & 44.5 & 20.6 & 1.02e-3 & 8.14e-4 & 6.41e-4 \\ \cline{3-10}
                &                           & Random & Lamarckian & 16.5 & 28.7 & 9.8 & 1.17e-3 & 1.01e-3 & 6.50e-4 \\ \cline{2-10}
                & \multirow{4}{*}{Xavier} & Xavier & Xavier & 15.9 & 25.4 & 7.3 & 1.14e-3 & 7.96e-4 & 4.87e-4 \\ \cline{3-10}
                &                           & Lamarckian & Xavier & 18.5 & 40.2 & 10.6 & 9.65e-4 & 6.49e-4 & 5.02e-4 \\ \cline{3-10}
                &                           & Lamarckian & Lamarckian & 17.6 & 37.5 & 14.0 & 1.00e-3 & 6.85e-4 & 4.74e-4 \\ \cline{3-10}
                &                           & Xavier & Lamarckian & 15.6 & 25.2 & 6.4 & 9.58e-4 & 6.99e-4 & \bf4.85e-4 \\ \cline{2-10}
                & \multirow{4}{*}{Kaiming} & Kaiming & Kaiming & 15.6 & 25.2 & 5.8 & 9.52e-4 & 7.08e-4 & 5.07e-4 \\ \cline{3-10}
                &                           & Lamarckian & Kaiming & 18.4 & 40.3 & 12.6 & \bf\emph{7.96e-4} & \bf\emph{6.29e-4} & 4.87e-4 \\ \cline{3-10}
                &                           & Lamarckian & Lamarckian & 18.4 & 40.4 & 12.6 & 9.99e-4 & 6.73e-4 & 4.87e-4 \\ \cline{3-10}
                &                           & Kaiming & Lamarckian & 15.5 & 23.5 & 7.1 & 1.13e-3 & 7.54e-4 & 5.56e-4 \\ \cline{1-10}
\multirow{12}{*}{80} & \multirow{4}{*}{Random} & Random & Random & 14.8 & 20.6 & 5.8 & 1.07e-3 & 9.53e-4 & 6.25e-4 \\ \cline{3-10}
                &                           & Lamarckian & Random & 20.9 & 48.0 & 12.5 & 8.96e-4 & 7.79e-4 & 6.29e-4 \\ \cline{3-10}
                &                           & Lamarckian & Lamarckian & 18.9 & 39.9 & 11.1 & 9.96e-4 & 7.99e-4 & 6.66e-4 \\ \cline{3-10}
                &                           & Random & Lamarckian & 14.8 & 20.6 & 5.2 & 1.17e-3 & 9.97e-4 & 8.65e-4 \\ \cline{2-10}
                & \multirow{4}{*}{Xavier} & Xavier & Xavier & 15.6 & 23.9 & 4.0 & 1.13e-3 & 9.07e-4 & 6.16e-4 \\ \cline{3-10}
                &                           & Lamarckian & Xavier & 17.0 & 31.9 & 6.8 & \bf8.03e-4 & \bf6.31e-4 & 5.02e-4 \\ \cline{3-10}
                &                           & Lamarckian & Lamarckian & 17.8 & 32.9 & 7.6 & 8.34e-4 & 6.55e-4 & 5.01e-4 \\ \cline{3-10}
                &                           & Xavier & Lamarckian & 15.3 & 22.5 & 4.5 & 1.11e-3 & 8.35e-4 & 5.60e-4 \\ \cline{2-10}
                & \multirow{4}{*}{Kaiming} & Kaiming & Kaiming & 14.9 & 22.1 & 3.8 & 1.09e-3 & 7.88e-4 & 5.12e-4 \\ \cline{3-10}
                &                           & Lamarckian & Kaiming & 17.4 & 32.6 & 8.8 & 8.53e-4 & 6.77e-4 & 5.00e-4 \\ \cline{3-10}
                &                           & Lamarckian & Lamarckian & 16.8 & 29.7 & 8.8 & 8.63e-4 & 6.58e-4 & 4.95e-4 \\ \cline{3-10}
                &                           & Kaiming & Lamarckian & 14.8 & 21.7 & 4.7 & 1.16e-3 & 8.49e-4 & \bf\emph{4.82e-4} \\ \cline{1-10}

        \hline
    \end{tabular}
    \caption{\label{table:coal_1} 
    Statistics of best genomes over 20 repeats on Coal dataset}

\end{table*}

\begin{table*}[p]
\centering
    \fontsize{7.5pt}{8.5pt} \selectfont
    \begin{tabular}{|C{28pt}|C{35pt}|C{48pt}|C{52pt}|C{30pt}|C{30pt}|C{30pt}|C{33pt}|C{33pt}|C{33pt}|} \hline
      BP    & Weight        & Crossover  & Mutation & Avg  & Avg   &  Avg      & Worst & Avg & Best\\
      Epochs & Initialize  & Inheritance  & New Edge/Node  & Node & Edge  & Rec Edge  & MAE   & MAE & MAE\\ \hline \hline
\multirow{12}{*}{1} & \multirow{4}{*}{Random} & Random & Random & 77.7 & 16.1 & 7.0 & 1.56e-2 & 1.40e-2 & 1.20e-2 \\ \cline{3-10}
&                           & Lamarckian & Random & 52.7 & 22.2 & 18.1 & 6.58e-3 & 5.68e-3 & 4.70e-3 \\ \cline{3-10}
&                           & Lamarckian & Lamarckian & 42.7 & 25.6 & 18.3 & 6.81e-3 & 5.52e-3 & 4.22e-3 \\ \cline{3-10}
&                           & Random & Lamarckian & 69.3 & 16.7 & 4.6 & 1.63e-2 & 1.36e-2 & 9.50e-3 \\ \cline{2-10}
& \multirow{4}{*}{Xavier} & Xavier & Xavier & 66.4 & 17.0 & 10.2 & 1.08e-2 & 8.82e-3 & 7.28e-3 \\ \cline{3-10}
&                           & Lamarckian & Xavier & 78.7 & 15.8 & 27.8 & 3.82e-3 & 3.32e-3 & {\bf2.64e-3} \\ \cline{3-10}
&                           & Lamarckian & Lamarckian & 68.8 & 14.9 & 30.4 & 3.90e-3 & 3.39e-3 & 2.72e-3 \\ \cline{3-10}
&                           & Xavier & Lamarckian & 56.1 & 26.0 & 4.8 & 7.57e-3 & 5.69e-3 & 4.38e-3 \\ \cline{2-10}
& \multirow{4}{*}{Kaiming} & Kaiming & Kaiming & 73.2 & 28.9 & 6.2 & 9.71e-3 & 8.00e-3 & 6.74e-3 \\ \cline{3-10}
&                           & Lamarckian & Kaiming & 81.7 & 16.4 & 39.1 & \bf3.73e-3 & {\bf3.19e-3} & 2.77e-3 \\ \cline{3-10}
&                           & Lamarckian & Lamarckian & 77.0 & 16.2 & 35.5 & \bf3.73e-3 & 3.34e-3 & 2.77e-3 \\ \cline{3-10}
&                           & Kaiming & Lamarckian & 56.0 & 17.1 & 4.0 & 6.66e-3 & 5.72e-3 & 3.67e-3 \\ \cline{1-10}
\multirow{12}{*}{5} & \multirow{4}{*}{Random} & Random & Random & 88.0 & 35.2 & 15.8 & 1.29e-2 & 1.12e-2 & 8.90e-3 \\ \cline{3-10}
&                           & Lamarckian & Random & 75.2 & 16.7 & 30.1 & 3.77e-3 & 3.42e-3 & 3.16e-3 \\ \cline{3-10}
&                           & Lamarckian & Lamarckian & 70.8 & 18.1 & 31.1 & 3.70e-3 & 3.36e-3 & 2.97e-3 \\ \cline{3-10}
&                           & Random & Lamarckian & 92.7 & 27.9 & 12.5 & 1.28e-2 & 1.12e-2 & 9.78e-3 \\ \cline{2-10}
& \multirow{4}{*}{Xavier} & Xavier & Xavier & 64.3 & 30.6 & 15.4 & 7.92e-3 & 6.82e-3 & 4.88e-3 \\ \cline{3-10}
&                           & Lamarckian & Xavier & 96.5 & 14.0 & 22.0 & 3.14e-3 & 2.85e-3 & 2.67e-3 \\ \cline{3-10}
&                           & Lamarckian & Lamarckian & 90.8 & 13.2 & 25.9 & 3.12e-3 & 2.85e-3 & 2.65e-3 \\ \cline{3-10}
&                           & Xavier & Lamarckian & 73.8 & 21.6 & 10.8 & 7.16e-3 & 5.97e-3 & 4.78e-3 \\ \cline{2-10}
& \multirow{4}{*}{Kaiming} & Kaiming & Kaiming & 85.8 & 33.6 & 9.8 & 7.72e-3 & 7.08e-3 & 5.33e-3 \\ \cline{3-10}
&                           & Lamarckian & Kaiming & 91.2 & 13.8 & 23.6 & 3.06e-3 & 2.79e-3 & 2.50e-3 \\ \cline{3-10}
&                           & Lamarckian & Lamarckian & 91.7 & 14.0 & 26.5 & \bf3.04e-3 & {\bf2.76e-3} & {\bf2.47e-3} \\ \cline{3-10}
&                           & Kaiming & Lamarckian & 89.3 & 32.2 & 6.8 & 6.72e-3 & 5.42e-3 & 3.64e-3 \\ \cline{1-10}
\multirow{12}{*}{10} & \multirow{4}{*}{Random} & Random & Random & 91.6 & 39.7 & 12.8 & 6.52e-3 & 6.16e-3 & 5.94e-3 \\ \cline{3-10}
&                           & Lamarckian & Random & 67.2 & 19.4 & 27.3 & 3.25e-3 & 3.11e-3 & 2.94e-3 \\ \cline{3-10}
&                           & Lamarckian & Lamarckian & 84.3 & 15.4 & 22.4 & 3.41e-3 & 3.06e-3 & 2.85e-3 \\ \cline{3-10}
&                           & Random & Lamarckian & 92.2 & 40.7 & 10.9 & 6.47e-3 & 6.18e-3 & 6.02e-3 \\ \cline{2-10}
& \multirow{4}{*}{Xavier} & Xavier & Xavier & 88.2 & 40.4 & 9.9 & 6.41e-3 & 5.83e-3 & 5.02e-3 \\ \cline{3-10}
&                           & Lamarckian & Xavier & 91.8 & 13.4 & 19.6 & 2.89e-3 & 2.74e-3 & 2.51e-3 \\ \cline{3-10}
&                           & Lamarckian & Lamarckian & 91.2 & 12.7 & 17.9 & 2.88e-3 & 2.74e-3 & {\bf2.48e-3} \\ \cline{3-10}
&                           & Xavier & Lamarckian & 92.3 & 43.9 & 9.8 & 6.29e-3 & 5.87e-3 & 4.80e-3 \\ \cline{2-10}
& \multirow{4}{*}{Kaiming} & Kaiming & Kaiming & 93.1 & 28.4 & 8.8 & 6.36e-3 & 5.76e-3 & 5.17e-3 \\ \cline{3-10}
&                           & Lamarckian & Kaiming & 92.2 & 13.8 & 21.3 & 2.87e-3 & 2.70e-3 & 2.53e-3 \\ \cline{3-10}
&                           & Lamarckian & Lamarckian & 91.7 & 13.3 & 17.5 & \bf2.80e-3 & {\bf2.65e-3} & 2.50e-3 \\ \cline{3-10}
&                           & Kaiming & Lamarckian & 92.3 & 31.6 & 7.6 & 1.10e-2 & 5.87e-3 & 4.65e-3 \\ \cline{1-10}
\multirow{12}{*}{20} & \multirow{4}{*}{Random} & Random & Random & 91.5 & 48.8 & 5.8 & 4.56e-3 & 4.44e-3 & 4.36e-3 \\ \cline{3-10}
&                           & Lamarckian & Random & 62.7 & 16.2 & 20.9 & 2.92e-3 & 2.85e-3 & 2.75e-3 \\ \cline{3-10}
&                           & Lamarckian & Lamarckian & 53.5 & 16.6 & 23.2 & 2.96e-3 & 2.83e-3 & 2.73e-3 \\ \cline{3-10}
&                           & Random & Lamarckian & 90.9 & 69.2 & 5.4 & 4.49e-3 & 4.41e-3 & 4.01e-3 \\ \cline{2-10}
& \multirow{4}{*}{Xavier} & Xavier & Xavier & 91.0 & 69.0 & 5.5 & 4.52e-3 & 4.42e-3 & 4.32e-3 \\ \cline{3-10}
&                           & Lamarckian & Xavier & 65.8 & 14.1 & 16.4 & 2.78e-3 & {\bf2.64e-3} & 2.52e-3 \\ \cline{3-10}
&                           & Lamarckian & Lamarckian & 86.7 & 12.8 & 16.6 & 2.82e-3 & {\bf2.63e-3} & {\bf\emph{2.45e-3}} \\ \cline{3-10}
&                           & Xavier & Lamarckian & 90.7 & 73.0 & 5.2 & 4.48e-3 & 4.38e-3 & 3.98e-3 \\ \cline{2-10}
& \multirow{4}{*}{Kaiming} & Kaiming & Kaiming & 91.5 & 52.5 & 4.2 & 4.51e-3 & 4.41e-3 & 4.12e-3 \\ \cline{3-10}
&                           & Lamarckian & Kaiming & 77.5 & 12.8 & 18.8 & 2.81e-3 & 2.66e-3 & 2.53e-3 \\ \cline{3-10}
&                           & Lamarckian & Lamarckian & 74.1 & 13.3 & 18.2 & \bf2.74e-3 & 2.67e-3 & 2.59e-3 \\ \cline{3-10}
&                           & Kaiming & Lamarckian & 91.4 & 52.5 & 5.4 & 4.51e-3 & 4.36e-3 & 3.80e-3 \\ \cline{1-10}
\multirow{12}{*}{40} & \multirow{4}{*}{Random} & Random & Random & 90.3 & 85.0 & 4.8 & 3.92e-3 & 3.89e-3 & 3.85e-3 \\ \cline{3-10}
&                           & Lamarckian & Random & 71.0 & 14.3 & 16.1 & 2.84e-3 & 2.75e-3 & 2.58e-3 \\ \cline{3-10}
&                           & Lamarckian & Lamarckian & 62.4 & 14.4 & 14.3 & 2.82e-3 & 2.74e-3 & 2.66e-3 \\ \cline{3-10}
&                           & Random & Lamarckian & 90.7 & 73.3 & 3.3 & 3.93e-3 & 3.89e-3 & 3.85e-3 \\ \cline{2-10}
& \multirow{4}{*}{Xavier} & Xavier & Xavier & 90.8 & 72.5 & 3.8 & 3.92e-3 & 3.86e-3 & 3.58e-3 \\ \cline{3-10}
&                           & Lamarckian & Xavier & 94.3 & 15.8 & 13.3 & 2.67e-3 & {\bf2.58e-3} & 2.48e-3 \\ \cline{3-10}
&                           & Lamarckian & Lamarckian & 89.9 & 20.3 & 14.2 & \bf2.66e-3 & 2.59e-3 & {\bf2.47e-3} \\ \cline{3-10}
&                           & Xavier & Lamarckian & 90.3 & 85.2 & 3.2 & 3.92e-3 & 3.85e-3 & 3.52e-3 \\ \cline{2-10}
& \multirow{4}{*}{Kaiming} & Kaiming & Kaiming & 90.7 & 76.7 & 3.2 & 3.94e-3 & 3.85e-3 & 3.57e-3 \\ \cline{3-10}
&                           & Lamarckian & Kaiming & 86.2 & 11.8 & 14.7 & 2.67e-3 & 2.59e-3 & {\bf2.47e-3} \\ \cline{3-10}
&                           & Lamarckian & Lamarckian & 86.2 & 12.1 & 15.2 & 2.67e-3 & {\bf\emph{2.57e-3}} & {\bf2.46e-3} \\ \cline{3-10}
&                           & Kaiming & Lamarckian & 90.9 & 64.1 & 2.2 & 3.91e-3 & 3.80e-3 & 3.49e-3 \\ \cline{1-10}
\multirow{12}{*}{80} & \multirow{4}{*}{Random} & Random & Random & 90.1 & 80.0 & 3.2 & 3.61e-3 & 3.59e-3 & 3.57e-3 \\ \cline{3-10}
&                           & Lamarckian & Random & 64.3 & 16.7 & 10.6 & 2.85e-3 & 2.71e-3 & 2.64e-3 \\ \cline{3-10}
&                           & Lamarckian & Lamarckian & 57.4 & 13.5 & 12.2 & 2.80e-3 & 2.68e-3 & 2.57e-3 \\ \cline{3-10}
&                           & Random & Lamarckian & 89.8 & 92.3 & 3.5 & 3.62e-3 & 3.60e-3 & 3.57e-3 \\ \cline{2-10}
& \multirow{4}{*}{Xavier} & Xavier & Xavier & 90.5 & 76.5 & 2.1 & 3.57e-3 & 3.43e-3 & 3.28e-3 \\ \cline{3-10}
&                           & Lamarckian & Xavier & 85.5 & 19.8 & 10.8 & 2.66e-3 & {\bf2.58e-3} & {\bf2.47e-3} \\ \cline{3-10}
&                           & Lamarckian & Lamarckian & 94.0 & 24.2 & 9.7 & \bf\emph{2.65e-3} & {\bf2.58e-3} & {\bf2.46e-3} \\ \cline{3-10}
&                           & Xavier & Lamarckian & 90.6 & 62.7 & 2.4 & 3.59e-3 & 3.40e-3 & 3.00e-3 \\ \cline{2-10}
& \multirow{4}{*}{Kaiming} & Kaiming & Kaiming & 90.6 & 72.0 & 1.4 & 3.48e-3 & 3.34e-3 & 3.18e-3 \\ \cline{3-10}
&                           & Lamarckian & Kaiming & 90.2 & 28.6 & 11.2 & 2.66e-3 & {\bf2.58e-3} & 2.54e-3 \\ \cline{3-10}
&                           & Lamarckian & Lamarckian & 90.3 & 11.4 & 8.8 & 2.67e-3 & {\bf2.58e-3} & 2.48e-3 \\ \cline{3-10}
&                           & Kaiming & Lamarckian & 90.5 & 76.2 & 1.9 & 3.53e-3 & 3.32e-3 & 2.93e-3 \\ \cline{1-10}
        \hline
    \end{tabular}
    \caption{\label{table:wind_1} Statistics of best genomes over 20 repeats on wind dataset}

\end{table*}

\begin{table*}
\centering
\begin{minipage}{0.48\textwidth}
\centering
    \small
    \begin{tabular}{|C{25pt}|C{28pt}|C{28pt}|C{28pt}|C{28pt}|C{28pt}|} \hline
      BP  & & & & & \\ 
      Epochs & Type & K-K-K & K-L-K & K-L-L  & K-K-L \\  \hline \hline 

\multirow{4}{*}{80} & K-K-K & /  & \bf0.0000  & \bf0.0000  & 0.1685   \\ \cline{2-6}
                    & K-L-K & \bf 0.0000  & /  & 0.4091  & \bf0.0000   \\ \cline{2-6}
                    & K-L-L & \bf 0.0000  & 0.4091  & /  & \bf0.0000   \\ \cline{2-6}
                    & K-K-L & 0.1685  & \bf0.0000  & \bf0.0000  & /   \\ \cline{1-6}
\multirow{4}{*}{40} & K-K-K & /  & \bf0.0000  & \bf0.0000  & \bf0.0234   \\ \cline{2-6}
                    & K-L-K & \bf 0.0000  & /  & 0.0860  & \bf0.0000   \\ \cline{2-6}
                    & K-L-L & \bf 0.0000  & 0.0860  & /  & \bf0.0000   \\ \cline{2-6}
                    & K-K-L & \bf 0.0234  & \bf0.0000  & \bf0.0000  & /   \\ \cline{1-6}
\multirow{4}{*}{20} & K-K-K & /  & \bf0.0000  & \bf0.0000  & 0.0903   \\ \cline{2-6}
                    & K-L-K & \bf 0.0000  & /  & 0.2989  & \bf0.0000   \\ \cline{2-6}
                    & K-L-L & \bf 0.0000  & 0.2989  & /  & \bf0.0000   \\ \cline{2-6}
                    & K-K-L & 0.0903  & \bf0.0000  & \bf0.0000  & /   \\ \cline{1-6}
\multirow{4}{*}{10} & K-K-K & /  & \bf0.0000  & \bf0.0000  & 0.1824   \\ \cline{2-6}
                    & K-L-K & \bf 0.0000  & /  & 0.0583  & \bf0.0000   \\ \cline{2-6}
                    & K-L-L & \bf 0.0000  & 0.0583  & /  & \bf0.0000   \\ \cline{2-6}
                    & K-K-L & 0.1824  & \bf0.0000  & \bf0.0000  & /   \\ \cline{1-6}
\multirow{4}{*}{5} & K-K-K & /  & \bf0.0000  & \bf0.0000  & \bf0.0000   \\ \cline{2-6}
                    & K-L-K & \bf 0.0000  & /  & 0.3180  & \bf0.0000   \\ \cline{2-6}
                    & K-L-L & \bf 0.0000  & 0.3180  & /  & \bf0.0000   \\ \cline{2-6}
                    & K-K-L & \bf 0.0000  & \bf0.0000  & \bf0.0000  & /   \\ \cline{1-6}
\multirow{4}{*}{1} & K-K-K & /  & \bf0.0000  & \bf0.0000  & \bf0.0000   \\ \cline{2-6}
                    & K-L-K & \bf 0.0000  & /  & 0.0632  & \bf0.0000   \\ \cline{2-6}
                    & K-L-L & \bf 0.0000  & 0.0632  & /  & \bf0.0000   \\ \cline{2-6}
                    & K-K-L & \bf 0.0000  & \bf0.0000  & \bf0.0000  & /   \\ \cline{1-6}
        \hline
    \end{tabular}
    \caption{\label{table:kaiming_wind} 
    Mann–Whitney U test $p$-values comparing Kaiming weight initialization and inheritance strategies for wind turbine dataset. $p$-values in bold indicate a statistically significant difference with $\alpha = 0.05$.}
\end{minipage}
\begin{minipage}{0.48\textwidth}
\centering
    \small
    \begin{tabular}{|C{25pt}|C{28pt}|C{28pt}|C{28pt}|C{28pt}|C{28pt}|} \hline
      BP  & & & & & \\ 
    Epochs & Type & X-X-X & X-L-X & X-L-L & X-X-L \\  \hline \hline 
    \multirow{4}{*}{80} & X-X-X & /  & \bf0.0000  & \bf0.0000  & 0.4409   \\ \cline{2-6}
                        & X-L-X & \bf 0.0000  & /  & 0.2625  & \bf0.0000   \\ \cline{2-6}
                        & X-L-L & \bf 0.0000  & 0.2625  & /  & \bf0.0000   \\ \cline{2-6}
                        & X-X-L & 0.4409  & \bf0.0000  & \bf0.0000  & /   \\ \cline{1-6}
    \multirow{4}{*}{40} & X-X-X & /  & \bf0.0000  & \bf0.0000  & 0.1143   \\ \cline{2-6}
                        & X-L-X & \bf 0.0000  & /  & 0.2896  & \bf0.0000   \\ \cline{2-6}
                        & X-L-L & \bf 0.0000  & 0.2896  & /  & \bf0.0000   \\ \cline{2-6}
                        & X-X-L & 0.1143  & \bf0.0000  & \bf0.0000  & /   \\ \cline{1-6}
    \multirow{4}{*}{20} & X-X-X & /  & \bf0.0000  & \bf0.0000  & 0.3575   \\ \cline{2-6}
                        & X-L-X & \bf 0.0000  & /  & 0.4623  & \bf0.0000   \\ \cline{2-6}
                        & X-L-L & \bf 0.0000  & 0.4623  & /  & \bf0.0000   \\ \cline{2-6}
                        & X-X-L & 0.3575  & \bf0.0000  & \bf0.0000  & /   \\ \cline{1-6}
    \multirow{4}{*}{10} & X-X-X & /  & \bf0.0000  & \bf0.0000  & 0.3676   \\ \cline{2-6}
                        & X-L-X & \bf 0.0000  & /  & 0.4730  & \bf0.0000   \\ \cline{2-6}
                        & X-L-L & \bf 0.0000  & 0.4730  & /  & \bf0.0000   \\ \cline{2-6}
                        & X-X-L & 0.3676  & \bf0.0000  & \bf0.0000  & /   \\ \cline{1-6}
    \multirow{4}{*}{5} & X-X-X & /  & \bf0.0000  & \bf0.0000  & \bf0.0012   \\ \cline{2-6}
                        & X-L-X & \bf 0.0000  & /  & 0.4946  & \bf0.0000   \\ \cline{2-6}
                        & X-L-L & \bf 0.0000  & 0.4946  & /  & \bf0.0000   \\ \cline{2-6}
                        & X-X-L & \bf 0.0012  & \bf0.0000  & \bf0.0000  & /   \\ \cline{1-6}
    \multirow{4}{*}{1} & X-X-X & /  & \bf0.0000  & \bf0.0000  & \bf0.0000   \\ \cline{2-6}
                        & X-L-X & \bf 0.0000  & /  & 0.2285  & \bf0.0000   \\ \cline{2-6}
                        & X-L-L & \bf 0.0000  & 0.2285  & /  & \bf0.0000   \\ \cline{2-6}
                        & X-X-L & \bf 0.0000  & \bf0.0000  & \bf0.0000  & /   \\ \cline{1-6}

        \hline
    \end{tabular}
    \caption{\label{table:xavier_wind} 
    Mann–Whitney U test $p$-values comparing Xavier weight initialization and inheritance strategies for wind turbine dataset. $p$-values in bold indicate a statistically significant difference with $\alpha = 0.05$.}
\end{minipage}
\vspace{4mm}

\begin{minipage}{0.48\textwidth}
\centering
    \small
    \begin{tabular}{|C{25pt}|C{28pt}|C{28pt}|C{28pt}|C{28pt}|C{28pt}|} \hline
      BP  & & & & & \\ 
      Epochs & Type & K-K-K & K-L-K & K-L-L  & K-K-L \\  \hline \hline 
    \multirow{4}{*}{1} & K-K-K & /  & 0.1897  & 0.1427  & 0.1971   \\ \cline{2-6}
                        & K-L-K & 0.1897  & /  & 0.2989  & 0.0599   \\ \cline{2-6}
                        & K-L-L & 0.1427  & 0.2989  & /  & \bf0.0249   \\ \cline{2-6}
                        & K-K-L & 0.1971  & 0.0599  & \bf0.0249  & /   \\ \cline{1-6}
    \multirow{4}{*}{5} & K-K-K & /  & 0.0599  & \bf0.0382  & 0.3084   \\ \cline{2-6}
                        & K-L-K & 0.0599  & /  & 0.4409  & 0.0818   \\ \cline{2-6}
                        & K-L-L & \bf 0.0382  & 0.4409  & /  & 0.0538   \\ \cline{2-6}
                        & K-K-L & 0.3084  & 0.0818  & 0.0538  & /   \\ \cline{1-6}
    \multirow{4}{*}{10} & K-K-K & /  & 0.1197  & 0.1143  & \bf0.0077   \\ \cline{2-6}
                        & K-L-K & 0.1197  & /  & 0.3180  & \bf0.0002   \\ \cline{2-6}
                        & K-L-L & 0.1143  & 0.3180  & /  & \bf0.0007   \\ \cline{2-6}
                        & K-K-L & \bf 0.0077  & \bf0.0002  & \bf0.0007  & /   \\ \cline{1-6}
    \multirow{4}{*}{20} & K-K-K & /  & \bf0.0077  & \bf0.0104  & 0.3676   \\ \cline{2-6}
                        & K-L-K & \bf 0.0077  & /  & 0.4946  & \bf0.0083   \\ \cline{2-6}
                        & K-L-L & \bf 0.0104  & 0.4946  & /  & \bf0.0045   \\ \cline{2-6}
                        & K-K-L & 0.3676  & \bf0.0083  & \bf0.0045  & /   \\ \cline{1-6}
    \multirow{4}{*}{40} & K-K-K & /  & \bf0.0360  & 0.1971  & 0.1824   \\ \cline{2-6}
                        & K-L-K & \bf 0.0360  & /  & 0.1824  & \bf0.0062   \\ \cline{2-6}
                        & K-L-L & 0.1971  & 0.1824  & /  & 0.0509   \\ \cline{2-6}
                        & K-K-L & 0.1824  & \bf0.0062  & 0.0509  & /   \\ \cline{1-6}
    \multirow{4}{*}{80} & K-K-K & /  & \bf0.0234  & \bf0.0169  & 0.1251   \\ \cline{2-6}
                        & K-L-K & \bf 0.0234  & /  & 0.3474  & \bf0.0003   \\ \cline{2-6}
                        & K-L-L & \bf 0.0169  & 0.3474  & /  & \bf0.0005   \\ \cline{2-6}
                        & K-K-L & 0.1251  & \bf0.0003  & \bf0.0005  & /   \\ \cline{1-6}

        \hline
    \end{tabular}
    \caption{\label{table:kaiming_coal} 
    Mann–Whitney U test $p$-values comparing Kaiming weight initialization and inheritance strategies for coal dataset. $p$-values in bold indicate a statistically significant difference with $\alpha = 0.05$.}
\end{minipage}
\begin{minipage}{0.48\textwidth}
\centering
    \small
    \begin{tabular}{|C{25pt}|C{28pt}|C{28pt}|C{28pt}|C{28pt}|C{28pt}|} \hline
      BP  & & & & & \\ 
    Epochs & Type & X-X-X & X-L-X & X-L-L & X-X-L \\  \hline \hline 
    \multirow{4}{*}{1} & X-X-X & /  & \bf0.0083  & \bf0.0036  & 0.3986   \\ \cline{2-6}
                        & X-L-X & \bf 0.0083  & /  & 0.4730  & \bf0.0022   \\ \cline{2-6}
                        & X-L-L & \bf 0.0036  & 0.4730  & /  & \bf0.0012   \\ \cline{2-6}
                        & X-X-L & 0.3986  & \bf0.0022  & \bf0.0012  & /   \\ \cline{1-6}
    \multirow{4}{*}{5} & X-X-X & /  & \bf0.0382  & \bf0.0158  & 0.2124   \\ \cline{2-6}
                        & X-L-X & \bf 0.0382  & /  & 0.1488  & \bf0.0104   \\ \cline{2-6}
                        & X-L-L & \bf 0.0158  & 0.1488  & /  & \bf0.0036   \\ \cline{2-6}
                        & X-X-L & 0.2124  & \bf0.0104  & \bf0.0036  & /   \\ \cline{1-6}
    \multirow{4}{*}{10} & X-X-X & /  & 0.1971  & 0.3474  & 0.1488   \\ \cline{2-6}
                        & X-L-X & 0.1971  & /  & 0.3779  & \bf0.0481   \\ \cline{2-6}
                        & X-L-L & 0.3474  & 0.3779  & /  & 0.0860   \\ \cline{2-6}
                        & X-X-L & 0.1488  & \bf0.0481  & 0.0860  & /   \\ \cline{1-6}
    \multirow{4}{*}{20} & X-X-X & /  & \bf0.0077  & \bf0.0249  & 0.1754   \\ \cline{2-6}
                        & X-L-X & \bf 0.0077  & /  & 0.1971  & 0.1092   \\ \cline{2-6}
                        & X-L-L & \bf 0.0249  & 0.1971  & /  & 0.2204   \\ \cline{2-6}
                        & X-X-L & 0.1754  & 0.1092  & 0.2204  & /   \\ \cline{1-6}
    \multirow{4}{*}{40} & X-X-X & /  & \bf0.0206  & 0.0538  & 0.0818   \\ \cline{2-6}
                        & X-L-X & \bf 0.0206  & /  & 0.2367  & 0.2124   \\ \cline{2-6}
                        & X-L-L & 0.0538  & 0.2367  & /  & 0.3882   \\ \cline{2-6}
                        & X-X-L & 0.0818  & 0.2124  & 0.3882  & /   \\ \cline{1-6}
    \multirow{4}{*}{80} & X-X-X & /  & \bf0.0000  & \bf0.0000  & 0.0538   \\ \cline{2-6}
                        & X-L-X & \bf 0.0000  & /  & 0.2804  & \bf0.0001   \\ \cline{2-6}
                        & X-L-L & \bf 0.0000  & 0.2804  & /  & \bf0.0002   \\ \cline{2-6}
                        & X-X-L & 0.0538  & \bf0.0001  & \bf0.0002  & /   \\ \cline{1-6}
        \hline
    \end{tabular}
    \caption{\label{table:xavier_coal} 
    Mann–Whitney U test $p$-values comparing Xavier weight initialization and inheritance strategies for coal dataset. $p$-values in bold indicate a statistically significant difference with $\alpha = 0.05$.}
\end{minipage}

\end{table*}

\begin{table*}
\centering

\begin{minipage}{0.48\textwidth}
\centering
    \small
    \begin{tabular}{|C{25pt}|C{28pt}|C{28pt}|C{28pt}|C{28pt}|C{28pt}|} \hline
      BP  & & & & & \\ 
    Epochs & Type & R-R-R & R-L-R & R-L-L & R-R-L \\ \hline \hline 
        \multirow{4}{*}{80} & R-R-R & /  & \bf0.0000  & \bf0.0000  & 0.0903   \\ \cline{2-6}
                            & R-L-R & \bf 0.0000  & /  & 0.1042  & \bf0.0000   \\ \cline{2-6}
                            & R-L-L & \bf 0.0000  & 0.1042  & /  & \bf0.0000   \\ \cline{2-6}
                            & R-R-L & 0.0903  & \bf0.0000  & \bf0.0000  & /   \\ \cline{1-6}
        \multirow{4}{*}{40} & R-R-R & /  & \bf0.0000  & \bf0.0000  & 0.3375   \\ \cline{2-6}
                            & R-L-R & \bf 0.0000  & /  & 0.2285  & \bf0.0000   \\ \cline{2-6}
                            & R-L-L & \bf 0.0000  & 0.2285  & /  & \bf0.0000   \\ \cline{2-6}
                            & R-R-L & 0.3375  & \bf0.0000  & \bf0.0000  & /   \\ \cline{1-6}
        \multirow{4}{*}{20} & R-R-R & /  & \bf0.0000  & \bf0.0000  & 0.2367   \\ \cline{2-6}
                            & R-L-R & \bf 0.0000  & /  & 0.1552  & \bf0.0000   \\ \cline{2-6}
                            & R-L-L & \bf 0.0000  & 0.1552  & /  & \bf0.0000   \\ \cline{2-6}
                            & R-R-L & 0.2367  & \bf0.0000  & \bf0.0000  & /   \\ \cline{1-6}
        \multirow{4}{*}{10} & R-R-R & /  & \bf0.0000  & \bf0.0000  & 0.2538   \\ \cline{2-6}
                            & R-L-R & \bf 0.0000  & /  & 0.0568  & \bf0.0000   \\ \cline{2-6}
                            & R-L-L & \bf 0.0000  & 0.0568  & /  & \bf0.0000   \\ \cline{2-6}
                            & R-R-L & 0.2538  & \bf0.0000  & \bf0.0000  & /   \\ \cline{1-6}
        \multirow{4}{*}{5} & R-R-R & /  & \bf0.0000  & \bf0.0000  & 0.4302   \\ \cline{2-6}
                            & R-L-R & \bf 0.0000  & /  & 0.2124  & \bf0.0000   \\ \cline{2-6}
                            & R-L-L & \bf 0.0000  & 0.2124  & /  & \bf0.0000   \\ \cline{2-6}
                            & R-R-L & 0.4302  & \bf0.0000  & \bf0.0000  & /   \\ \cline{1-6}
        \multirow{4}{*}{1} & R-R-R & /  & \bf0.0000  & \bf0.0000  & 0.3474   \\ \cline{2-6}
                            & R-L-R & \bf 0.0000  & /  & 0.1684  & \bf0.0000   \\ \cline{2-6}
                            & R-L-L & \bf 0.0000  & 0.1684  & /  & \bf0.0000   \\ \cline{2-6}
                            & R-R-L & 0.3474  & \bf0.0000  & \bf0.0000  & /   \\ \cline{1-6}
        \hline
    \end{tabular}
    \caption{\label{table:random_wind} 
    Mann–Whitney U test $p$-values comparing uniform random weight initialize and inheritance strategies for wind turbine dataset. $p$-values in bold indicate a statistically significant difference with $\alpha = 0.05$.}
\end{minipage}
\begin{minipage}{0.48\textwidth}
\centering
    \small
    \begin{tabular}{|C{25pt}|C{28pt}|C{28pt}|C{28pt}|C{28pt}|C{28pt}|} \hline
      BP  & & & & & \\ 
    Epochs & Type & R-R-R & R-L-R & R-L-L & R-R-L \\  \hline \hline 
    \multirow{4}{*}{1} & R-R-R & /  & \bf0.0000  & \bf0.0000  & 0.3676   \\ \cline{2-6}
                        & R-L-R & \bf 0.0000  & /  & 0.4409  & \bf0.0000   \\ \cline{2-6}
                        & R-L-L & \bf 0.0000  & 0.4409  & /  & \bf0.0000   \\ \cline{2-6}
                        & R-R-L & 0.3676  & \bf0.0000  & \bf0.0000  & /   \\ \cline{1-6}
    \multirow{4}{*}{5} & R-R-R & /  & \bf0.0030  & \bf0.0015  & 0.0632   \\ \cline{2-6}
                        & R-L-R & \bf 0.0030  & /  & 0.2538  & \bf0.0000   \\ \cline{2-6}
                        & R-L-L & \bf 0.0015  & 0.2538  & /  & \bf0.0000   \\ \cline{2-6}
                        & R-R-L & 0.0632  & \bf0.0000  & \bf0.0000  & /   \\ \cline{1-6}
    \multirow{4}{*}{10} & R-R-R & /  & \bf0.0014  & \bf0.0001  & 0.3575   \\ \cline{2-6}
                        & R-L-R & \bf 0.0014  & /  & \bf0.0283  & \bf0.0011   \\ \cline{2-6}
                        & R-L-L & \bf 0.0001  & \bf0.0283  & /  & \bf0.0000   \\ \cline{2-6}
                        & R-R-L & 0.3575  & \bf0.0011  & \bf0.0000  & /   \\ \cline{1-6}
    \multirow{4}{*}{20} & R-R-R & /  & \bf0.0002  & \bf0.0003  & 0.4091   \\ \cline{2-6}
                        & R-L-R & \bf 0.0002  & /  & 0.2714  & \bf0.0002   \\ \cline{2-6}
                        & R-L-L & \bf 0.0003  & 0.2714  & /  & \bf0.0003   \\ \cline{2-6}
                        & R-R-L & 0.4091  & \bf0.0002  & \bf0.0003  & /   \\ \cline{1-6}
    \multirow{4}{*}{40} & R-R-R & /  & \bf0.0002  & \bf0.0011  & 0.2047   \\ \cline{2-6}
                        & R-L-R & \bf 0.0002  & /  & 0.3676  & \bf0.0000   \\ \cline{2-6}
                        & R-L-L & \bf 0.0011  & 0.3676  & /  & \bf0.0001   \\ \cline{2-6}
                        & R-R-L & 0.2047  & \bf0.0000  & \bf0.0001  & /   \\ \cline{1-6}
    \multirow{4}{*}{80} & R-R-R & /  & \bf0.0000  & \bf0.0001  & 0.1617   \\ \cline{2-6}
                        & R-L-R & \bf 0.0000  & /  & 0.2538  & \bf0.0000   \\ \cline{2-6}
                        & R-L-L & \bf 0.0001  & 0.2538  & /  & \bf0.0000   \\ \cline{2-6}
                        & R-R-L & 0.1617  & \bf0.0000  & \bf0.0000  & /   \\ \cline{1-6}

        \hline
    \end{tabular}
    \caption{\label{table:random_coal} 
    Mann–Whitney U test $p$-values comparing uniform random weight initialize and inheritance strategies for coal dataset. $p$-values in bold indicate a statistically significant difference with $\alpha = 0.05$.}
\end{minipage}
\end{table*}

This work utilized two real world data sets for predicting time series data with RNNs\footnote{These data sets are publicly available at EXAMM repository: https://github.com/travisdesell/exact/tree/master/datasets/}. The first data set comes from data collected from 12 burners of a coal-fired power plant, the second data set is wind turbine engine data from 2013 to 2020, collected and made available by ENGIE's La Haute Borne open data windfarm\footnote{https://opendata-renewables.engie.com}. Both datasets are multivariate (with 12 and 88 parameters, respectively), non-seasonal, and the parameter recordings are not independent. Furthermore, they are very long. The power plant data consists of 10-day worth of per-minute data while the wind turbine data consists of readings every 10 minutes from 2013 to 2020. \emph{Main flame intensity} was chosen as the output parameter for the coal dataset and \emph{average active power} was selected as output parameter for wind turbine data set.

\paragraph{Hyperparameter Settings}

Each EXAMM run used $10$ islands, each with a maximum capacity of $10$ genomes. New RNNs were generated via mutation at a rate of 70\%, intra-island crossover at a rate of 20\%, and inter-island crossover at a rate of 10\%. $10$ out of EXAMM's $11$ mutation operations were utilized (all except for \emph{split edge}), and each was chosen with a uniform 10\% chance. EXAMM generated new nodes by selecting from simple neurons, $\Delta$-RNN, GRU, LSTM, MGU, and UGRNN memory cells uniformly at random. Recurrent connections could span any time-skip generated randomly between $\mathcal{U}(1,10)$. Backpropagation (BP) through time was run with a learning rate of $\eta = 0.001$ and used Nesterov momentum with $\mu = 0.9$. For the memory cells with forget gates, the forget gate bias had a value of $1.0$ added to it (motivated by \cite{jozefowicz2015empirical}).  To prevent exploding gradients, gradient scaling~\cite{pascanu2013difficulty} was used when the norm of the gradient exceeded a threshold of $1.0$. To combat vanishing gradients, gradient boosting (the opposite of scaling) was used when the gradient norm was below $0.05$. These parameters have been selected by hand-tuning during prior experience with these data sets.

\paragraph{Experimental Design}

Our hypotheses were that {\it i)}, utilizing Lamarckian weight initialization would provide performance improvements over uniform random, Xavier and Kaiming inheritance, and {\it ii)}, it could potentially allow for networks to be effectively evolved using less BP epochs. To provide a comprehensive exploration, we set up experiments where the initial networks initialized weights with the uniform random, Xavier and Kaiming strategies (as the Lamarckian strategies could not yet be used). In these experiments, we tested the combinations of the two Lamarckian strategies with the initial weight inheritance strategy. Note that when Lamarckian weight inheritance is used for crossover and Xavier for weight inheritance, this is identical to the Lamarckain strategy used by~\cite{prellberg2018lamarckian}, so this strategy is investigated as well.

BP epochs of $1$, $5$, $10$, $20$, $40$ and $80$ per genome generated were examined. To have a fair comparison between the test cases, the total number of BP epochs for each search was set to $200k$ for each test. This resulted in the total number of genomes generated during evolution for the tests being $200k$, $40k$, $20k$, $10k$, $5k$ and $2.5k$ respectively.  In total for each initial weight strategy (uniform random, Kaiming and Xavier) there were 24 different experiments done with different BP epochs and strategies for crossover and mutation weight initialization for both the coal fired power plant and wind turbine datasets. All the experiments were repeated $20$ times allowing the Mann–Whitney U-test for statistical significance to be used.

\paragraph{Computing Environment}
Results were gathered using Rochester Institute of Technology's research computing systems. This system consists of 2304 Intel® Xeon® Gold 6150 CPU 2.70GHz cores and 24 TB RAM, with compute nodes running the RedHat Enterprise Linux 7 system. All the experiments utilized 72 cores.

\paragraph{Results}

Figure~\ref{fig:box_plot} presents box plots of the best genome fitness from 20 repeats of each of the experiments using Xavier, Kaiming, and uniform random weight initialization and their combinations with the Lamarckian strategies. These results are summarized in Tables~\ref{table:coal_1} and~\ref{table:wind_1} which presents the best, average, and worst global best genome mean average error (MAE) at the end of the 20 repeated tests for each experiment performed on coal and wind datasets. The best performing experiments are highlighted in bold for each number of BP epochs, and the overall best experiment is highlighted in bold and italics. 

In the average case, for both datasets, all the best performing genomes were found using Lamarckian weight inheritance on crossover, further in the best cases all but 3 also utilized Lamarckian weight inheritance for crossover, however in these cases they still utilized Lamarckian weight inheritance for mutation. For any number of BP epochs, using purely Xavier or Kaiming initialization never performed the best. As additional validation to the results and prior work, Xavier generally performed better than Kaiming, because the RNN nodes and memory cells used the symmetric activation function \emph{tanh} which Xavier has been shown to perform better on~\cite{glorot2010understanding}.

Further strengthening these results, Tables~\ref{table:kaiming_wind},~\ref{table:xavier_wind},~\ref{table:kaiming_coal},~\ref{table:xavier_coal},~\ref{table:random_wind}, and~\ref{table:random_coal} present Mann-Whitney U tests of statistical significance comparing the varying weight initialization tests against each other for each number of BP epochs. Strategies are labeled by \emph{initial genome strategy}-\emph{crossover strategy}-\emph{mutation strategy}, so \eg, \emph{K-L-K} would use Kaiming for the initial genomes, Lamarckian on crossover operations, and Kaiming for components generated by mutation. For the wind turbine data, we see very strong statistical significance in most cases, highlighting that the improvements from the Lamarckian strategy. While the statistical significance is less strong in some cases on the coal dataset, interestingly the statistical significance increases with the number of BP epochs utilized perhaps due to the fact more training time enable quicker convergence to local or global minima.

For the wind turbine dataset, utilizing Lamarckian weight inheritance also shows the ability to significantly reduce the number of BP epochs required for training, which in turn allows for more time to be spent evolving the RNN architectures. Using the Lamarckian strategies, the overall best results in the average case were found using 40 BP epochs, and the overall best result was found at 20 BP epochs. Alternately, Kaiming and Xavier found their best average case and best overall results at 80 BP epochs. This trend was less clear with the coal data, where Kaiming and Xavier found their  best results with less BP epochs with the average cases being 40 and 10 BP epochs respectively, and overall best at 5 and 50 BP epochs, respectively. Here, Lamarckian strategies found their best results at 40 BP epochs on the average case, and 10 BP epochs for the overall best. However, the Lamarckian strategies still performed the best here overall, perhaps suggesting that this dataset required more architectural evolution to find good results as the best networks tended to have more nodes and edges.

\section{Conclusions}



This work is an experimental study on the effects of weight initialization and weight inheritance in neuroevolution. It compares the well known Kaiming and Xavier weight initialization strategies to two Lamarckian weight inheritance strategies, once based on recombining parental weights during crossover, and another using statistical information of parental weights to assign new weights in mutation operations.  This is done in the context of the Evolutionary eXploration of Augmenting Memory Models (EXAMM) neuroevolution algorithm, which progressively evolves and trains RNNs for time series data prediction using a direct encoding strategy. Experiments were done using large scale real world time series data sets, one generated from a coal fired power plant and another from a wind turbine.

A comprehensive suite of tests was run, finding with statistical significance that the Lamarckian strategies outperform Xavier and Kaiming weight initialization for generating new RNNs through EXAMM. Further, these Lamarckian strategies are also shown to be able to reduce the number of backpropagation epochs required to train the generated neural networks, allowing the neuroevolution algorithm to be able to perform more architectural evolution.  These results validate a commonly held view that Lamarckian weight inheritance strategies can improve the performance of neuroevolution algorithms~\cite{prellberg2018lamarckian,desell2018accelerating,ororbia2019examm}, which to the authors knowledge has not been rigorously compared to state-of-the-art Xavier and Kaiming weight initialization.

These results present a strong case for the use of Lamarckian weight inheritance strategies for neuroevolution of recurrent neural networks for time series data prediction. The Lamarckian strategies presented are generic and can be applied to any direct encoding neuroevolution algorithm. Future work will expand these results to convolutional neural networks as well as recurrent neural networks used for natural language processing tasks (which tend to have wider but shallower architectures). Given these results as motivation, investigating new Lamarckian strategies to further enhance performance will also be done.




\section{ Acknowledgments}
Most of the computation of this research was done on the high performance computing clusters of Research Computing at Rochester Institute of Technology~\cite{https://doi.org/10.34788/0s3g-qd15}. We would like to thank the Research Computing team for their assistance and the support they generously offered to ensure that the heavy computation this study required was available.

\clearpage
\pagebreak
\newpage

\bibliography{references.bib}

\end{document}